\documentclass[runningheads]{llncs}
\usepackage{graphicx}
\usepackage{comment}
\usepackage{eucal}
\usepackage{amsmath}
\usepackage{amssymb}
\usepackage{tikz}
\usepackage{chngpage}
\usepackage{bibnames}
\usepackage{subcaption}
\usepackage{setspace}
\usepackage{paralist}

\usepackage{hyperref}

\begin{document}
\def\checkmark{\tikz\fill[scale=0.4](0,.35) -- (.25,0) -- (1,.7) -- (.25,.15) -- cycle;}

\title{Faster-LTN: a neuro-symbolic, end-to-end object detection architecture}

\author{Francesco Manigrasso\inst{1}\orcidID{0000-0002-4151-8880} \and
Filomeno Davide Miro\inst{1} \and
Lia Morra\inst{1}\orcidID{0000-0003-2122-7178}\and
Fabrizio Lamberti\inst{1}\orcidID{0000-0001-7703-1372}}
\authorrunning{F. Manigrasso et al.}
\institute{Politecnico di Torino, Dipartimento di Automatica e Informatica, Torino, Italy 
\email{francesco.manigrasso@polito.it, filomenodavide.miro@studenti.polito.it, \{lia.morra, fabrizio.lamberti\}@polito.it}\\
}
%\and ABC Institute, Rupert-Karls-University Heidelberg, Heidelberg, Germany\\
\maketitle              % typeset the header of the contribution
\begin{abstract}
The detection of semantic relationships between objects represented in an image is one of the fundamental challenges in image interpretation.
Neural-Symbolic techniques, such as Logic Tensor Networks (LTNs), allow the combination of semantic knowledge representation and reasoning with the ability to efficiently learn from examples typical of neural networks. We here propose Faster-LTN, an object detector composed of a convolutional backbone and an LTN. To the best of our knowledge, this is the first attempt to combine both frameworks in an end-to-end training setting. This architecture is trained by optimizing a grounded theory which combines labelled examples with prior knowledge, in the form of logical axioms.  Experimental comparisons show competitive performance with respect to the traditional Faster R-CNN architecture.

\end{abstract}

\section{Introduction}
A long-standing problem in Semantic Image Interpretation (SII) and related tasks is how to combine learning from data with existing background knowledge in the form of relational knowledge or logical axioms \cite{2019arXiv190609954A}. Neural-Symbolic (NeSy) integration, which aims at integrating symbolic knowledge representation and learning with machine learning techniques \cite{de2020statistical}, can provide an elegant and principled solution to augment state-of-the-art deep neural networks with these novel capabilities, increasing their performance, robustness and explainability. 

The present work leverages the Logic Tensor Network (LTN) paradigm that was proposed by Serafini, Donadello and d'Avila Garcez \cite{Donadello2017LogicTN,badreddine2020logic}. In very simple terms, LTNs operate by interpreting (or \textit{grounding}) a First-Order Logic (FOL) as functions on real vectors, which parameters can be trained via stochastic gradient descents to maximize the satisfiability of a given theory.  LTNs have been successfully applied to the tasks of part-of relationship detection \cite{Donadello2017LogicTN} and visual relationship detection \cite{Donadello2019CompensatingSI}. Previous works have shown how LTNs can compensate the lack of supervision (e.g., in few-shot learning scenarios) by relying on logical axioms  derived from pre-existing knowledge bases. 

To close the semantic gap between the symbolic (concept) and subsymbolic (pixel) levels, LTNs for SII rely on convolutional neural networks (CNNs) to extract semantic features which form the basis for grounding object instances in a real vector. Previous works \cite{Donadello2019CompensatingSI,Donadello2017LogicTN}
relied on pre-trained CNNs, which however suffer from all the limitations traditionally associated with deep learning, namely, the need for a large-scale annotated dataset for training, and lack of interpretability. To fully reap the benefits of NeSy techniques in SII, end-to-end architectures in which the LTN is jointly trained with the feature extraction CNN are needed. 

In this work, we propose Faster-LTN, an object detector which unifies the Faster R-CNN object detector with a LTN-based classification head. Differently from previous works \cite{Donadello2017LogicTN,Donadello2019CompensatingSI}, both modules are jointly trained in an end-to-end fashion. The logical constraints imposed by the LTN can thus shape the training of the convolutional layers, that are no longer purely data-driven. To achieve this objective, we propose several modifications to the original LTN formulation to increase the architecture scalability and deal with data imbalance. Experimental results on the PASCAL VOC and PASCAL PART datasets show that Faster-LTN converges to competitive performance with respect to purely neural architectures, thus proving the feasibility of this approach. The Faster-LTN was implemented in Keras and is available at \url{https://gitlab.com/grains2/Faster-LTN}.

The rest of the paper is organized as follows. In Section~\ref{sec:Relatedwork}, related work is presented. In Section~\ref{sec:methods}, different variations of the Faster-LTN architecture are presented, after a brief introduction to the theory behind LTNs. Section~\ref{sec:Experiments} presents the experimental setting and results. Finally, conclusions are drawn.

\section{Related work}
\label{sec:Relatedwork}
A natural image is comprised of scenes, objects and parts, all interconnected by a complex network of spatial and semantic relationships. Thus, developing semantic image interpretation (SII) components requires to recognize a hierarchy of components, and entails both robust visual perception and the ability to encode and (reason about) visual relationships. Several techniques have been proposed to augment Convolutional Neural Networks (CNNs) with relationship representation and reasoning capabilities, including Relational Network~\cite{shanahan2020explicitly}, Graph Neural Networks~\cite{lamb2020graph} and Neural-Symbolic (NeSy) techniques~\cite{yi2018neural,Donadello2017LogicTN,Donadello2019CompensatingSI}. For a more general introduction to NeSy techniques, the reader is referred to recent surveys~\cite{besold2017neural,garcez2019neural}.

Many recent approaches extract features from CNNs to a subsequent symbolic or neuro-symbolic module~\cite{10.1007/978-3-319-10605-2_27,Donadello2017LogicTN,Donadello2019CompensatingSI,10.1007/978-3-319-46448-0_51}. Yuke Zhu et al.~\cite{10.1007/978-3-319-10605-2_27} use a Markov Logic Network (MLN) to process text information with associated visual features; a knowledge base is used to represent relations between objects using visual, physical, and categorical attributes. Kenneth Marino et al.~\cite{Marino2017TheMY} incorporate a Graph Search Neural Network (GSNN) into a classification network. Donatello et al.~\cite{Donadello2017LogicTN} and  Cewu Lu et al.~\cite{10.1007/978-3-319-46448-0_51}  have demonstrated the use of visual features to train LTNs for visual relationship detection, in form of \textit{subject-verb-object} triplets or \textit{part of} relationships. These works demonstrate how NeSy techniques enable the definition of logical axioms that serve as high-level inductive biases, driving the network to find the optimal solution that is compatible with said inductive biases. However, since in the above-mentioned cases the feature extraction and the classification networks are trained separately, the CNN cannot leverage these additional inductive biases during training. 

There are, however, some practical hurdles associated with the training of NeSy architectures. Scalability, when dealing with large amounts of data, is a known issue associated with symbolic AI~\cite{Krieken2020AnalyzingDF}.  For this reason, many NeSy architectures rely on a conventional object detector to provide an initial list of candidate objects~\cite{Donadello2017LogicTN}, thus disregarding the effect of the background and simplifying (i.e., reducing) the scale of the problem. In this work, we compare several strategies that are effectively capable of training a LTN-based object detector from scratch, taking into account the effect of the background and the resulting data imbalance.

Another aspect related to scalability is the choice of aggregation function and fuzzy logic operators. Emilie van Krieken et al.~\cite{Krieken2020AnalyzingDF} and Samy Badreddine~\cite{badreddine2020logic} found substantial differences between   differential fuzzy logic operators in terms of computational efficiency, scalability, gradients, and ability to handle exceptions, which are important characteristics in a learning setting. Their analysis lays the groundwork for the present FasterLTN architecture, which incorporates and extends the log-product aggregator analyzed in~\cite{Krieken2020AnalyzingDF}.

\section{The Faster-LTN architecture}
\label{sec:methods}

\begin{figure}[tb]
\centering
	\includegraphics[width=\textwidth]{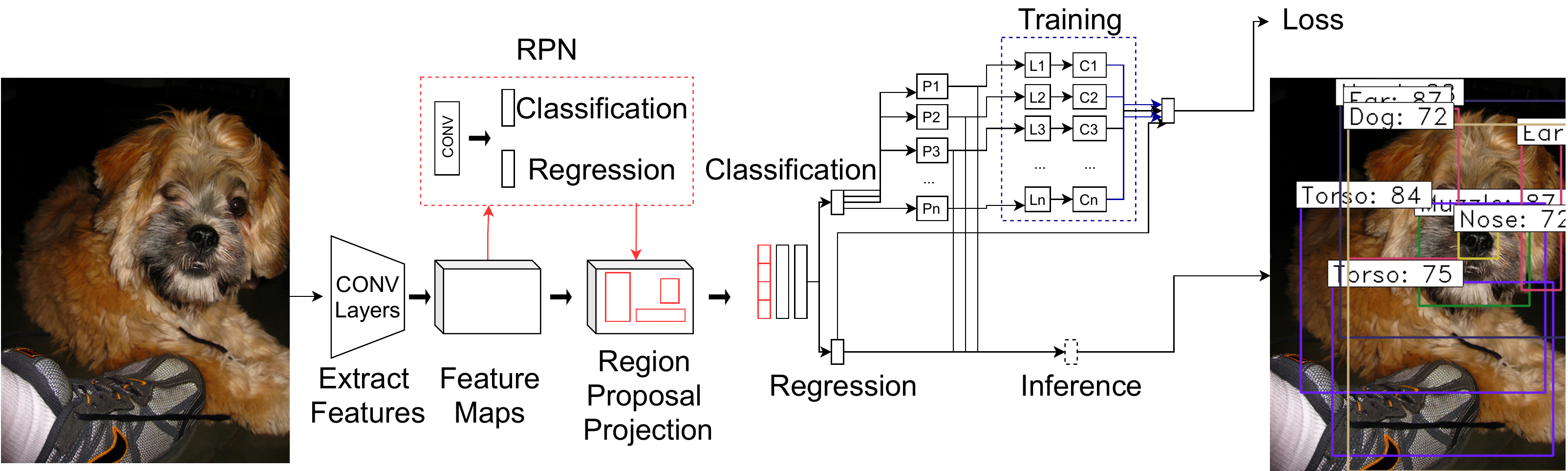}
	\caption{Faster-LTN architecture. The first part of the architecture, up to the RPN, is the same as in the Faster R-CNN network~\cite{Ren2015FasterRT}. The feature maps associated to the RPN proposals are extracted by the backbone, concatenated and passed to the LTN, which includes a collection of predicates $P_i$, each corresponding to a specific class. At training time, a batch of labelled examples in the training dataset are used to define a partial theory $\mathcal{T}_{expl}$. Each positive or negative example corresponds to a positive or negative literal (L) for the corresponding predicates. The truth value of the aggregated clauses (C) is maximized to find the optimal grounding $\mathcal{G}^{*}$. At inference time, the truth value of the predicates $P_i$ is computed. 
} 
	\label{Figure:Faster-LTN}
\end{figure}

This section describes the Faster-LTN architecture and training procedure in detail.  
An overview of the overall architecture is presented in Figure~\ref{Figure:Faster-LTN}. We first summarize the Faster R-CNN overall architecture (Section~\ref{sec:fasterrcnn}). Then, we introduce the main concepts behind LTNs (Section~\ref{sec:ltn}) and their application to object detection (Section~\ref{sec:ltn_object}), referring the reader to~\cite{Donadello2017LogicTN,badreddine2020logic} for additional details.   Finally, the joint training procedure of Faster-LTN is explained in Section~\ref{sec:Faster-LTN}, highlighting the main changes introduced to make end-to-end training feasible.

\subsubsection{Architecture}

\subsection{Faster R-CNN}
\label{sec:fasterrcnn}

Faster R-CNN is a two-stage object detector composed of a Region Proposal Network (RPN) and a classification network with a shared backbone~\cite{Ren2015FasterRT}. For each anchor, the RPN generates a binary classification label (Background vs. foreground), while a regression layer computes the bounding box coordinates. Regions of Interest (ROIs) selected by the RPN are fed to an ROI Pooling layer, which extracts and resizes each proposal bounding box's features from the shared backbone. Feature maps of equal size are passed to the classifier. The classifier comprises two convolutional heads, a classification layer (with softmax activation) that computes the final object classification and a regression layer (with linear activation) that computes the bounding box.

Training of the RPN and classifier heads is performed jointly in an alternating fashion. At each forward pass (corresponding to one image), the RPN is trained and updated; then, the RPN output is kept fixed, and the detector head is updated. A fixed number of positive (object) and negative (Background) examples are selected at each step to train the classifier head. 

The loss is as a combination of regression and classification loss:
\begin{equation}
    L(\{p_i\}, \{b_i\}) = \frac{1}{n_c}\sum_{i}L_{cls}(p_i, p'_{i}) +  \lambda\frac{1}{n_r}\sum_{i}{p_i * L_{reg}(b_i, b'_{i})}
\end{equation}

In the Faster-LTN, we keep the RPN module intact and substitute the classifier head with an LTN.

\subsection{Logic Tensor Network} 
\label{sec:ltn}

\subsubsection{Grounding}
In the LTN framework, it is possible to encode a FOL language $\mathcal{L}$  by defining its interpretation domain as a subset of  $\mathbb{R}^{n}$.  In the LTN formalism, this process is called \textit{grounding}. 

Given the vector space $\mathbb{R}^{n}$, a grounding $\mathcal{G}$ for $\mathcal{L}$ has the following properties:
\begin{enumerate}
    \item $\mathcal{G}(c) \in \mathbb{R}^n$, for every $c \in \mathcal{C}$;
    \item $\mathcal{G}(P) \in \mathbb{R}^{n*k} \rightarrow [0,1]$, for every $p \in \mathcal{P}$
\end{enumerate}

The grounding of a set of \textbf{closed terms}  $t_{1},..,t_{m}$ of $\mathcal{L}$ in an atomic formula is defined as:
\begin{align}
 \mathcal{G} \left( \mathcal{P}\left( t_{1},...t_{m} \right)\right) =\mathcal{G}\left( P\right) \left( \mathcal{G}  \left( t_{1}\right),...,\mathcal{G}  \left( t_{m}\right)\right)	
\end{align}

Formulas can be connected with fuzzy logic \textit{operators} such as conjunctions ($\land$), disjunctions ($\lor $), and implications ($\implies$), including logical quantifiers ($\forall$ and $\exists$). Several real-valued, differentiable implementations are available in the fuzzy logic domain~\cite{Krieken2020AnalyzingDF}. Our implementation, as in~\cite{Donadello2017LogicTN}, is based on the Łukasiewicz~\cite{10.1007/978-3-642-36039-8_18} formulation: 
\begin{align}
	& \mathcal{G} \left( \neg \phi \right) = 1 - \mathcal{G}\left( \phi \right) \\
	& \mathcal{G} \left( \phi \vee \psi \right) = min(1,\mathcal{G}\left( \phi \right) + \mathcal{G}\left( \psi \right)) 
	%& \mathcal{G} \left( \phi \wedge \psi \right) = max(0, \mathcal{G}\left( \phi \right) + \mathbb{G}\left( \psi \right) -1 ) \\
	%& \mathcal{G} \left( \phi \rightarrow \psi \right) = min(1,1 - \mathbb{G}\left( \phi \right) + \mathcal{G}\left( \psi \right)) \\
%	& \label{eq:meanterm} \mathcal{G} \left( \forall x \phi(x) \right) = \lim_{T \to term(\mathcal{L}) } mean_{p}(\mathcal{G}(\phi(x)) | t \in T) 
	\end{align}

Predicate symbols are interpreted as functions that map real vectors to the interval $[0, 1]$, which can be interpreted as the predicate's degree of truth. A typical example is the \textit{is-a} predicate, which quantifies the existence of a given object. For instance, if $b=\mathcal{G}(x)$ is the grounding of a dog bounding box, than $\mathsf{\mathcal{G}(Dog)(v)}\simeq 1$. A logical constraint expressed in FOL allows to define its properties, i.e., $\forall x \left( \mathsf{Dog}(x)  \rightarrow \mathsf{hasMuzzle} \left( x\right) \right)$. 

In LTNs, predicates are typically defined as the generalization of the neural tensor network: 

	\begin{align}
	\label{eqn:predicate}
	\mathcal{G}\left( \mathcal{P} \right)( \mathbf{v}) = \sigma\left(\mathit{u_{P}^{T}}\tanh\left( \mathbf{v_{T}} W_{P}^{[1:k]} \mathbf{v} + V_{P} \mathbf{v} + \mathit{b_{p}} \right) \right)
	\end{align}
	where $\sigma$ is the sigmoid function, $W[1:k] \in \mathbb{R}^{k \times mn \times mn}$, $V_{p} \in \mathbb{R}^{k \times mn}$ ,$u_{p} \in \mathbb{R}^{k}$ and $ b_{p} \in \mathbb{R} $  are learnable tensors of parameters. With this formulation, the truth value of a clause can be determined by a neural
network which first computes the grounding of the \textit{literals} (i.e., atomic objects) contained in the clause, and then combines them using fuzzy logical operators, as defined by Eqs. 3-4. 
	
%The predicates' parameters are learned through gradient descent by solving the best satisfiability problem, as defined in the following.

\subsubsection{Grounded theory} A Grounded Theory (GT) $\mathcal{T}$ is defined by a pair $\langle \mathcal{K},\hat{\mathcal{G}} \rangle$, where the knowledge base  $\mathcal{K}$ is a set of closed formulas,  and  $\hat{\mathcal{G}}$ is a partial grounding. $\mathcal{K}$ is constructed from labelled examples, as well as logical axioms, as defined in Section ~\ref{sec:ltn_object}. In practice, a partial grounding is optimized since, qualitatively, our set $\mathcal{K}$ represents a limited and finite set of examples. A grounding $G$ \textbf{satisfies} a GT $\langle \mathcal{K},\hat{\mathcal{G}} \rangle$ if $\mathcal{G}$ completes $\hat{\mathcal{G}}$ and $\mathcal{G}\left(\phi\right) = 1 \ \forall \ \phi \in \mathcal{K}$.

\subsubsection{Best satisfability problem}
Given a grounding $\hat{\mathcal{G}}_{\theta}$,  where $\theta$ is the set of parameters of all predicates, the learning problem in LTNs is framed as a \textit{best satisfability problem} which consists in determining the values of $\Theta^{*}$ that maximize the truth values of the conjunction of all clauses $\ \phi \in \mathcal{K}$:

\begin{align}
\label{eq:best_sat_2}
\Theta^{*}= argmax_{\Theta} \hat{\mathcal{G}}_{\theta}\left(\bigwedge_{\phi \in \mathcal{K}} \phi \right) - \lambda|| \Theta ||_{2}^{2}
\end{align}
where $ \lambda|| \Theta ||_{2}^{2}$ is a regularization term.
In practical problems, it is unlikely that a grounded theory can be satisfiable in the classical sense. Hence, we opt instead to find the grounding which achieves the best possible satisfaction, while accounting for the inevitable exception to the rule.  Such exceptions can easily arise in the visual domain not only to account to allow the occasional deviation from the norm, but also to account for properties that are not visible. For instance, a cat has (usually) a tail, but a few cats may be tail-less; more frequently, the tail will be occluded or cut from the image.

\subsection{LTN for object detection}
\label{sec:ltn_object}

\subsubsection{A grounded theory for object detection}

Let us consider a set of bounding boxes $b \in \mathcal{B}$ with known class $c \in \mathcal{C}$. An object with bounding box $b_{n}$ is grounded by the vector:
\begin{align}
\label{eq:grounding_object}
\mathbf{v_{b_{n}}}= <\mathbf{z}_{b_{n}},b_{n} >  
\end{align}
Where $\mathbf{z}_{b_{n}} = f(I,b_{n})$ is an embedding feature vector, calculated by a convolutional neural network $f$, given an image $I$ and the bounding box coordinates $b_{n}$ predicted by the RPN layer. This is slightly different from previous works~\cite{Donadello2017LogicTN}, where the grounding of a bounding box was defined by the probability vector predicted by a pre-trained Faster R-CNN, and allows to effectively connect the convolutional layers and the LTN. 

We set the embedding $f(I,b_{n})$ to the output of the last fully connected layer of the classifier head, without softmax activation. Other choices are possible, e.g., by sum pooling the output of an earlier convolutional layer. 

The \textit{is-a} predicate for class $c \in \mathcal{C}$ is grounded by a tensor network, defined as in Eq. ~\ref{eqn:predicate}, which implements a one-vs-all classifier. It must be noticed that, differently from~\cite{Donadello2017LogicTN}, the \textit{is-a} predicate takes as input only the embedding features  $\mathbf{z}_{b_{n}}$, excluding the bounding box coordinates. This allows to retain one of the basic properties of object detectors, i.e., invariance to translation.

The \textit{part-of} predicate is defined over pairs of bounding boxes~\cite{Donadello2017LogicTN}. A pair of two generic bounding boxes $b_{m}$ and  $b_{l}$ is grounded by the vector: 
\begin{align}
\mathbf{v_{b_{m, l}}}= <\mathbf{z}_{b_{m}},b_{m},\mathbf{z}_{b_{l}},b_{l},ir_{m,l}>
\end{align}
where $ir_{m,l}$ is the \textit{containment ratio} defined as:

 \begin{equation}
   \label{eq:containmentratio}
   	ir_{m,l}= \frac{Area \left( b_{m}  \cap  b_{l} \right)}{Area \left( b_{m} \right)} 
  \end{equation}

The grounding $\mathcal{G}\left( \mathsf{part-of} \right)( \mathbf{v_{b_{m, l}}})$ is a neural tensor network as in Eq.~\ref{eqn:predicate}. 

\subsubsection{Defining a theory from labelled examples}
Let us now consider how a GT is constructed to solve the best satisfiability problem defined in Eq. ~\ref{eq:best_sat_2} for object detection. As in~\cite{Donadello2017LogicTN}, two grounded theories $\mathcal{T}_{expl}$ and $\mathcal{T}_{prior}$ are defined. The former, $\mathcal{T}_{expl}$, aggregates all the clauses derived from the labelled training set, essentially replicating the classical learning-by-example setting. The theory $\mathcal{T}_{prior}$, on the contrary, introduces \textit{logical} and \textit{mereological} constraints that represent prior knowledge or, in a more general sense, desirable properties of the final solution. 

In this work, two types of constraints are defined. First, we enforce \textit{mutual exclusion} through the clause:
\begin{align}
\label{eq:implication}
\forall x( P_1(x) \implies ( \neg P_2(x) \land ... \land \neg P_n(x)))
\end{align}
Eq. ~\ref{eq:implication} is translated into  $K(K-1))/2$ clauses, corresponding to all unordered class pairs over $K$ classes, e.g., ${\mathsf{Cat}(x) \implies \neg \mathsf{Person}(x)}$. %,  to indicate that an object belonging to class $c_i$ cannot belong to class $c_j$. 

Secondly, we impose \textit{mereological constraints} on the grounding of \textit{part-of} and \textit{is-a} predicates derived from an existing ontology (e.g., Wordnet) which includes  \textit{meronimy} (i.e., \textit{part-whole}) relationships. Axioms are included to specify that a \textit{part} cannot include another \textit{part}, that a \textit{whole} object cannot include another \textit{whole} object, and that each \textit{whole} is generally associated with a set of given \textit{parts}. An example of such axioms is as follows:
\begin{align}
\forall x,y \left(\mathsf{Cat}(x) \wedge \mathsf{partOf}(y,x) \to \mathsf{Tail}\left(y\right) \vee \mathsf{Head}\left(y\right) ... \vee \mathsf{Eye}\left(y\right)\right)
\label{eq:parts}
\end{align}
to indicate that if an object $y$ is classified as part of $x$ and $x$ is a cat, than $y$ can be only an object that we know is a part of the whole cat.  
Mereological constraints were enforced exploiting the KB developed in~\cite{Donadello2017LogicTN}, to which the reader is referred for further information.

\subsection{Faster-LTN}
\label{sec:Faster-LTN}

The overall architecture, illustrated in Figure ~\ref{Figure:Faster-LTN}, is an end-to-end system connecting a convolutional object detector with an LTN. 
Specifically,  the classifier head is modified, by removing the softmax activation, and feeding the output to the LTN. At training time, a GT is constructed as defined in Section ~\ref{section:training}. The LTN is implemented by defining three additional layers: \textit{Predicate}, \textit{Literal} and \textit{Clause} layers. For each class $c$, the corresponding literal computes the truth value of all positive (i.e., belonging to class $c$) and negative (i.e., not belonging to class $c$) examples. The Clause layer aggregates all literals for a given class, using the selected aggregation function. Additionally, it is possible to define clauses (e.g., for \textit{part-of} predicates) that take as input multiple literals. For the sake of simplicity, in Figure ~\ref{Figure:Faster-LTN} only $\mathcal{T}_{expl}$ is shown. The final loss of the LTN is given by summing $L_{LTN}$ with the regression loss, as for the RPN layer.

\subsubsection{Training}
\label{section:training}
In order to deal with memory constraints, a partial $\mathcal{T}_{expl}$ needs to be rebuilt with every batch of examples.  In the original implementation ~\cite{Donadello2017LogicTN}, the LTN was trained on the predictions of a pre-trained object detector, allowing for a relatively large batch size. In our setting, the LTN is trained on all proposals extracted by the RPN, and a separate batch is constructed for each image, taking into account background as well as foreground examples. It is worth noticing that one-vs-all classification amplifies the data imbalance between positive and negative examples for each class, even when the training batch consists of an equal number of objects and background proposals. 

\subsubsection{Aggregation function}

The chosen aggregator function is the log-product, which was shown in~\cite{Krieken2020AnalyzingDF} to scale well with the number of inputs, and which formulation is equivalent to the cross-entropy loss. However, in our case, this choice does not weight adequately the contribution of positive examples, given the high level of class imbalance. Hence, inspired by~\cite{8237586}, we introduce the focal log-product aggregation defined as:
\begin{equation}
\label{eq:log_sum}
L_{LTN} = -\sum_{j = 0}^{K}\sum_{i = 0}^{N} \alpha_c (1 - x_{i,j})^{\gamma}log\left(x_{i,j}\right)	\end{equation}
where $\alpha_c$ is a class-dependent weight factor, $\gamma$ enhances the contribution of literals with low truth value (i.e., misclassified examples),  $x_i$ is the literal of the $i$-th ROI in the  $j$-th class, $K$ is the number of classes and $N$ is the batch size.

To set the value of $\alpha_c$, we simply observe that for each training batch and each class $c$, the number of negative examples is given by the number of background examples (which is fixed during training), plus the positive examples that belong to other classes. Hence, we set $\alpha_{c} = \frac{1 - \beta}{1 - \beta^{pos_c}}$ and  $\alpha_c = \frac{1 - \beta}{1 - \beta^{neg_c}}$, for positive and negative examples respectively. Let $p(c)$ be the fraction of bounding boxes in the training set belonging to class $c$. Then, for a given batch the percentage of positive and negative examples becomes ${pos_c = \frac{N}{2}  p\left( c\right)}$ and $ neg_c = \frac{N}{2} + \frac{N}{2}  \left(1 - p\left( c\right)\right) $, respectively.

\section{Experiments}
\label{sec:Experiments}

\subsection{Dataset}
\label{subsec:DATASET}

Experiments were performed on the PASCAL VOC 2010~\cite{pascal-voc-2010} and PASCAL PART \cite{chen2014detect} benchmarks. For the latter, we selected 20 classes for whole objects and 39 classes for parts. All experiments are conducted on the trainval partition with 80:20 split. For PASCAL PART (10K images), we further experiment reducing the training set by 50\% by random selection: the number of images is thus roughly 8K for PASCAL PART and 4K for PASCAL PART REDUCED.

\subsection{Experimental setup}

\subsubsection{Faster R-CNN}
The architecture of the Faster R-CNN follows quite closely the original implementation ~\cite{Ren2015FasterRT}.  The backbone architecture was ResNet50 pretrained on ImageNet; the anchor scales were set to $128^{2}$,$256^{2}$, and $512^{2}$, with aspect ratios of 1:1, 1:2,and 2:1.
The number of RPN proposals is set to 300. For training the classifier head, 128 bounding boxes were randomly selected, with a ratio of 32:96  positive and negative examples, for the PASCAL VOC dataset; for PASCAL PART, 32 bounding boxes with 16:16 ratio. The network was trained for 100 epochs with the Adam optimizer; the learning rate was set to $10^{-5}$ for the first 60 epochs, and then reduced to $10^{-6}$. Regularization techniques included data augmentation (horizontal flip) and weight decay (with rate $5 \times 10^{-4}$). 

\subsubsection{Faster-LTN}
The architecture of Faster-LTN was the same as Faster R-CNN, except for the classifier head in which the LTN was embedded. 

Each predicate is defined by Eq.~\ref{eqn:predicate}, with $k=6$ kernels. Łukasiewicz's t--norm was chosen to encode the literals' disjunction, and the focal log-product, with $\gamma = 2$, was selected as the aggregation function.  $\mathcal{T}_{prior}$ included mutual exclusion constraints for PASCAL VOC, and mutual exclusion and mereological constraints for PASCAL PART experiments. In the latter case, the LTN was expanded to include \textit{part-of} predicates, but for the sake of comparison with Faster R-CNN, only the object detection performance was evaluated.

On the PASCAL VOC dataset, different experiments were performed with variations of the focal log-product aggregation function: with and without class weights $\alpha$, and with and without adding an additional predicate \textit{bg} to represent the background class. The experiments are denoted as Faster-LTN, Faster-LTN $\alpha$, Faster-LTN \textit{bg}, and Faster-LTN \textit{bg}$+\alpha$. Experiments on PASCAL-PART were performed with the Faster-LTN \textit{bg} configuration. All networks were trained for 150 epochs using the Adam optimizer,  with weight decay (decay rate $5 \times 10^{-4}$), random horizontal flip and L2 regularization ($\lambda$ is set to $5 \times 10^{-4}$.). The learning rate was set to $10^{-5}$ for the first 60 epochs, and then reduced to $10^{-6}$.

All  experiments were performed on the HPC@Polito cluster, equipped with V100 NVIDIA GPU. The performance metric was the mean Average Precision (MAP) implemented as in the PASCAL VOC challenge 2010 ~\cite{8594067}.

\begin{table}[tb]\tiny
	\centering

		\begin{tabular}{|c|c|c|c|c|c|c|}
			\hline
			% v1
			%alpha v2
			% bg v3
			%bg with alpha v4
			
			\textbf{Class} & \textbf{FR-CNN}& \textbf{FR-CNN FL} & \textbf{F-LTN} & \textbf{F-LTN  $\alpha$} & \textbf{F-LTN bg} & \textbf{F-LTN bg$+\alpha$} \\
			\hline
			aeroplane & 66.5	&	56.9	& 87.1			& 85.1			& 87.8		& 85.2
\\ \hline
			bicycle & 69.9	&	64.1	& 75.6		& 77.3		& 77.8		& 77.4
\\ \hline
			bird  & 70.8	&	68.4	&  84.9		&87.8			& 87.2		& 87.1
\\\hline
			boat  & 41.3&	35.8		& 59.7			& 70.3		&  62.2			& 67.1
\\\hline
			bottle  & 51.0	& 44.1	& 48.2			& 45.8		& 43.7		& 47.0
\\\hline
			bus  & 75.8	&	71.3	& 79.1		& 79.0 & 79.8			& 78.6
\\\hline
			car  & 59.0	&	53.1	& 60.0		& 58.7	& 62.9			&  60.1
\\\hline
			cat  & 92.4	& 90.0	&  93.5			& 92.4		& 94.1		& 94.8
\\\hline
			chair  & 32.1 &	32.7		&  53.4		&  42.8		 & 53.4		& 42.9
\\\hline
			cow  & 64.6	&60.7	& 67.1		& 66.3	& 60.1			& 72.6
\\\hline
			diningtable  & 57.2	&51.1	& 74.2			&  77.0		& 71.3			& 77.1
\\\hline
			dog  & 85.3	&	83.3	& 93.6		&92.3	& 92.5			& 92.0
\\\hline
			horse  & 61.1 &	62.3		& 82.2		& 80.4			& 85.4			& 85.0
\\\hline
			motorbike  & 62.0&	65.3		& 86.7			& 81.0			& 85.6			& 85.0
\\\hline
			person   &70.7	& 68.7		& 72.6		& 49.5		& 74.1		& 53.3
\\\hline
			pottedplant  &  29.0&	25.4		& 53.1		& 49.2			&  48.8		& 51.8
\\\hline
			sheep  & 62.2	&	62.1	&71.2		&71.4			& 74.7		& 69.1
\\\hline
			sofa  & 59.9	&	51.9	& 79.2			& 82.0			& 86.4			& 80.1
\\\hline
			train  & 73.3	& 73.2	& 75.4		& 77.2			& 79.6		& 81.6
\\\hline
			tvmonitor  & 68.7	& 63.3	& 78.5			& 76.6			& 77.1		& 76.6

\\\hline	\textbf{mAP } &\textbf{62.6}&	\textbf{59.2}		&\textbf{73.8}		& \textbf{72.1}		& \textbf{73.3}			& \textbf{73.25} \\
			\hline
		\end{tabular}

		\caption{Results of the Faster R-CNN (FR-CNN), Faster R-CNN with focal loss (FR-CNN FL), and Faster-LTN (F-LTN) on PASCAL VOC. }
			\label{table:map_pascal_voc}
		
	\end{table}

\subsection{Results}
Experiments on Pascal VOC, summarized in Table~\ref{table:map_pascal_voc}, show that Faster LTN achieved competitive and even superior results compared to the original Faster R-CNN architecture, with the mAP increasing from 62.6 to 73.8. In this version of the LTN, the only axiomatic constraint was the one imposing mutual exclusivity (see Eq.~\ref{eq:parts}).  We observed comparable performance when including the background as an additional class (mAP from 73.8 to 73.4); on the other hand, weighting positive and negative samples according to their frequency did not improve results (mAP from 73.8 to 72.1). 

Qualitatively, we observed that Faster LTN was able to detect more objects than Faster R-CNN. Given that log-product aggregation is mathematically equivalent to the cross-entropy loss, and the backbone is the same, this difference can be attributed to the different classification setting ($K$ one-vs-all classifiers instead of a single multi-class classifier) or the use of the focal loss \cite{8237586}. However, when changing the loss of the Faster R-CNN classifier head to the focal loss, performance dropped from 62.6 to 59.2. Hence, we attribute Faster-LTN performance to the greater flexibility offered by a more complex classifier head, with higher number of parameters. In fairness, Faster LTN took a few more epochs to reach convergence. 

\begin{figure}[!tb]

\minipage{0.5\textwidth}
  \includegraphics[width=\linewidth]{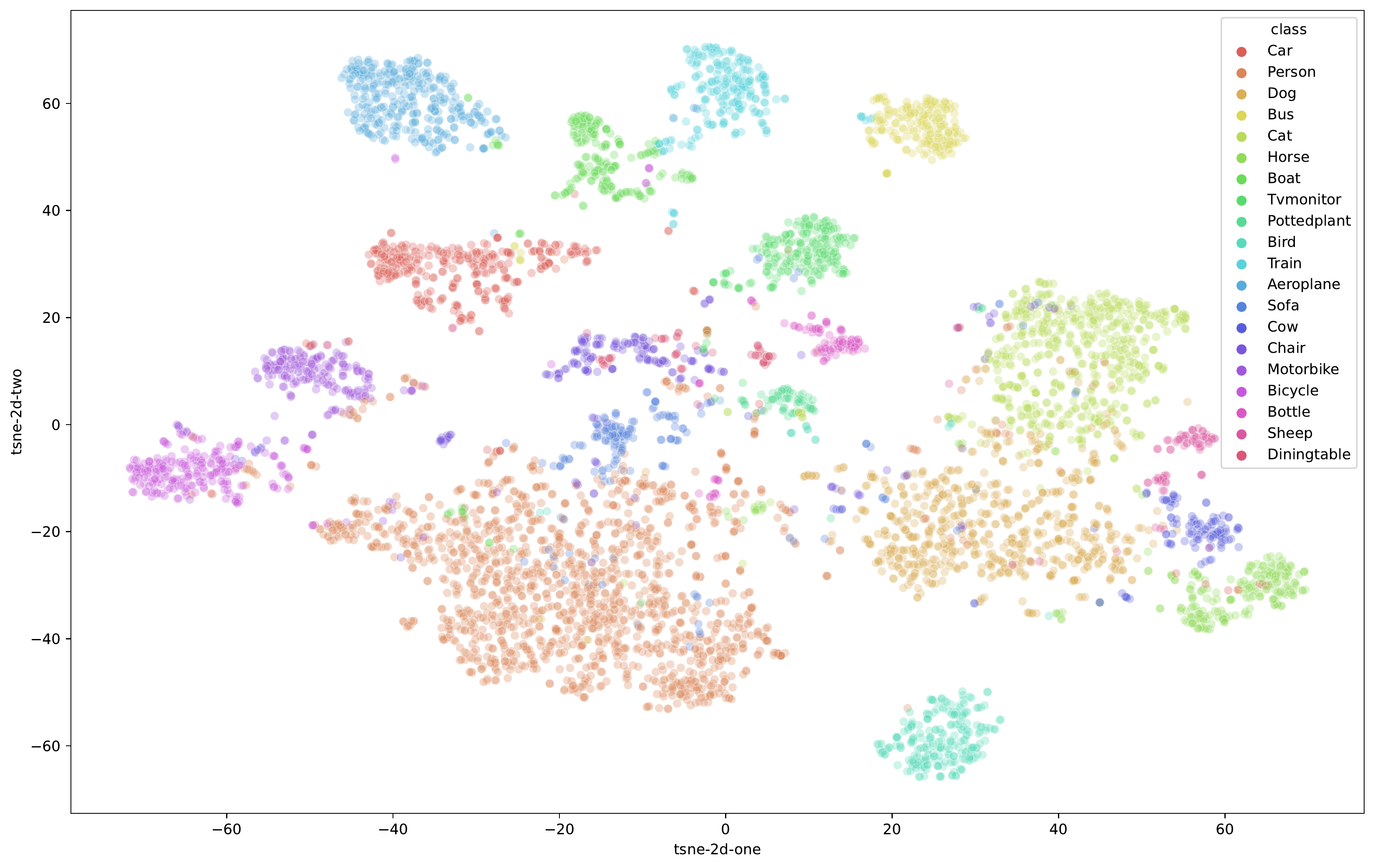}
  %\label{fig:Faster R-CNN}
\endminipage
\minipage{0.5\textwidth}
  \includegraphics[width=\textwidth]{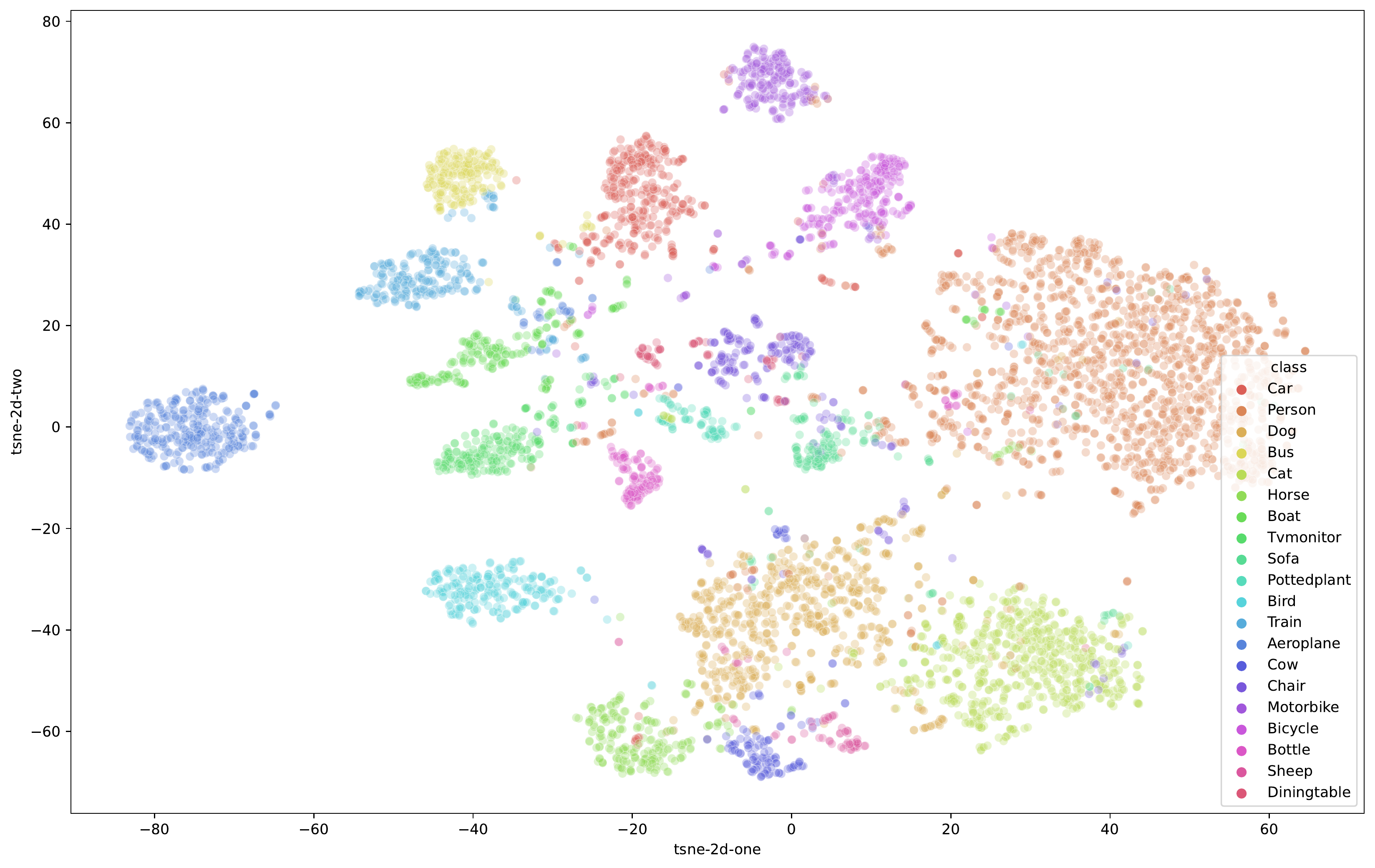}
  %\caption{Faster-LTN}\label{fig:Faster-LTN}
\endminipage\hfill

%\minipage{0.5\textwidth}
%  \includegraphics[width=\linewidth]{images/review/FASTERCNN_PART_HALF.pdf}
  %\label{fig:Faster R-CNN}
%\endminipage
%\minipage{0.5\textwidth}
%  \includegraphics[width=\textwidth]{images/review/FASTERLTN_PART_HALF.pdf}
  %\caption{Faster-LTN}\label{fig:Faster-LTN}
%\endminipage\hfill

%\minipage{0.5\textwidth}
  %\includegraphics[width=\linewidth]{images/fasterltnknowledge/feature_tsnet_100.pdf}/feature_tsnet_50.png}
%  \caption{Faster-LTN with knowledge}\label{fig:Faster-LTNknowledge}
%\endminipage\hfill
%\minipage{0.5\textwidth}
 % \includegraphics[width=\linewidth]{images/Faster-LTNpartof/feature_tsnet_50.png}
 % \caption{Faster-LTN with PartOF}\label{fig:Faster-LTNpartof}
%\endminipage\hfill
\caption{Comparison of the t-SNE embeddings of the features extracted for the \textit{whole} objects classes in the test test. Features extracted from Faster R-CNN (left) and Faster-LTN with axiomatic constraints (right). }
\label{t-sneplot}

\end{figure}

In the PASCAL PART experiments, shown in Table~\ref{table:map_pascal_part}, additional mereological axioms were included in $\mathcal{T}_{prior}$. This allowed to increase performance from 35.1 to 41.2; when reducing the training set size by half, the performance gap was maintained (28.5 to 32.8). The
comparable quality of the learned features is further supported by the t-SNE embeddings of the extracted features, which are shown in Figure~\ref{t-sneplot}.

% Faster-LTN bg mAP=40.1
\begin{table}[!tb]\tiny
	\centering

		\begin{tabular}{|c|c|c|c|}
			\hline

			Dataset &Metric & FR-CNN & F-LTN $\mathcal{T}_{prior} $  \\
			\hline
	
			PASCAL PART & mAP & 35.1		&  41.2	\\ \hline
			PASCAL PART REDUCED & mAP & 28.5	 & 32.8 \\ \hline

		\end{tabular}
			\caption{Comparison of Faster R-CNN and Faster-LTN (including mereological constraints) on the PASCAL PART dataset.}
			\label{table:map_pascal_part}
		
	\end{table}

\section{Conclusion and future works}
The availability of large scale, high quality, labelled datasets is one of the major hurdles in the application of deep learning. A tighter integration between perception and reasoning, which is enabled by emerging Neural-Symbolic techniques, allows to complement learning by examples with the integration of axiomatic background knowledge. 
In this paper, we introduced the Faster-LTN architecture, an end-to-end object detector composed by a convolutional backbone and RPN (based on the Faster R-CNN architecture) and a LTN module. The detector is trained end-to-end by maximizing the satisfiability of a grounded theory combining clauses derived from labelled examples with axiomatic constraints. 

Our goal was to establish the feasibility of this approach, and indeed the results, albeit preliminary, prove that Faster-LTN is competitive or can even outperform the baseline Faster R-CNN. However, the scalability of this approach to larger training sets and other object detector (e.g., single-stage detectors) should be further investigated. 
Through the Faster-LTN model, available at \url{https://gitlab.com/grains2/Faster-LTN}, we aim to provide a baseline architecture on which new experiments and applications can be built. Future work will investigate how high-level symbolic constraints can shape the learning process, increasing robustness in the presence of noise and dataset bias. 

\section*{Acknolewdgement}
The authors wish to thank Ivan Donadello for the helpful discussions. 
Computational resources were in part provided by HPC@POLITO, a project of Academic Computing at Politecnico di Torino (\url{http://www.hpc.polito.it}).

\bibliographystyle{splncs05}

\bibliography{main}

\end{document}